\documentclass[conference]{IEEEtran}
\IEEEoverridecommandlockouts
\usepackage{cite}
\usepackage{amsmath,amssymb,amsfonts}
\usepackage{algorithmic}
\usepackage{graphicx}
\usepackage{textcomp}
\usepackage{xcolor}
\usepackage{graphicx}
\usepackage{amsmath}
\usepackage[cmintegrals]{newtxmath}
\usepackage{mathtools}
\usepackage{xcolor}
\usepackage{tabularx}
\usepackage{newtxmath}
\usepackage{subcaption}
\usepackage{balance}

\begin{document}	
	\title{\huge Adversarial Attacks Against Deep Reinforcement Learning Framework in Internet of Vehicles}	
	\author{\IEEEauthorblockN{Anum Talpur and Mohan Gurusamy}
		\IEEEauthorblockA{\textit{\textit{Department of Electrical and Computer Engineering,}}\\ National University of Singapore, Singapore \\
			anum.talpur@u.nus.edu; gmohan@nus.edu.sg
		}}

\maketitle

\begin{abstract}
Machine learning (ML) has made incredible impacts and transformations in a wide range of vehicular applications. As the use of ML in Internet of Vehicles (IoV) continues to advance, adversarial threats and their impact have become an important subject of research worth exploring. In this paper, we focus on Sybil-based adversarial threats against a deep reinforcement learning (DRL)-assisted IoV framework and more specifically, DRL-based dynamic service placement in IoV. We carry out an experimental study with real vehicle trajectories to analyze the impact on service delay and resource congestion under different attack scenarios for the DRL-based dynamic service placement application. We further investigate the impact of the proportion of Sybil-attacked vehicles in the network. The results demonstrate that the performance is significantly affected by Sybil-based data poisoning attacks when compared to adversary-free healthy network scenario.
\end{abstract}

\begin{IEEEkeywords}
Sybil attack, deep reinforcement learning, service placement, multi-access edge computing, internet of vehicles.
\end{IEEEkeywords}

\IEEEpeerreviewmaketitle

\vspace{-10pt}

\section{Introduction}
\IEEEPARstart{I}{nternet} of Vehicles (IoV) witness promising development in vehicular designs by globally extending the accessibility and availability. The broader connectivity and wider interoperability of devices require modern control methods. Thus, the research evolves to incorporate artificial intelligence (AI) and machine learning (ML) into IoV applications \cite{IOVML2}. Driven by the dynamic nature of vehicles, deep reinforcement learning (DRL) has been a breakthrough technique for interactive and continual decision-making in IoVs. The use of DRL and its variants are widely explored in the literature to provide solutions toward motion planning and control, resource sharing, service placement, scheduling, security and many other aspects of vehicular networks \cite{driving,anumvtc,talpur2021machine}.\par 
Technological advances and innovation always bring in opportunities as well as limitations. The adversarial ML attack is one such limitation which is becoming an active area of research as ML grows to play a crucial role in a number of applications. The research predicts that 30\% of all cyberattacks will be adversarial attacks by 2022 \cite{Gartner}. According to one survey, Microsoft shares that 90\% of businesses don't have enough tools or techniques to secure their ML-operated systems \cite{microsoft}. The fact that ML is used to secure several other systems and frameworks, the security of the ML model itself has received little attention. Such attacks, their impact, their detection and prevention in the context of IoV applications are not much explored in the literature. With the advent of vehicle automation which involves the use of ML and its variants to perform services like motion planning and vehicle control, an adversarial attack may result in long delay and resource congestion leading to dire consequences of catastrophe in vehicles and danger to human lives. In this paper, we focus on the adversarial attacks on DRL framework in IoVs. \par 
\begin{figure}[htbp]
	\begin{center}
		\includegraphics[width=2.7in,height=2in]{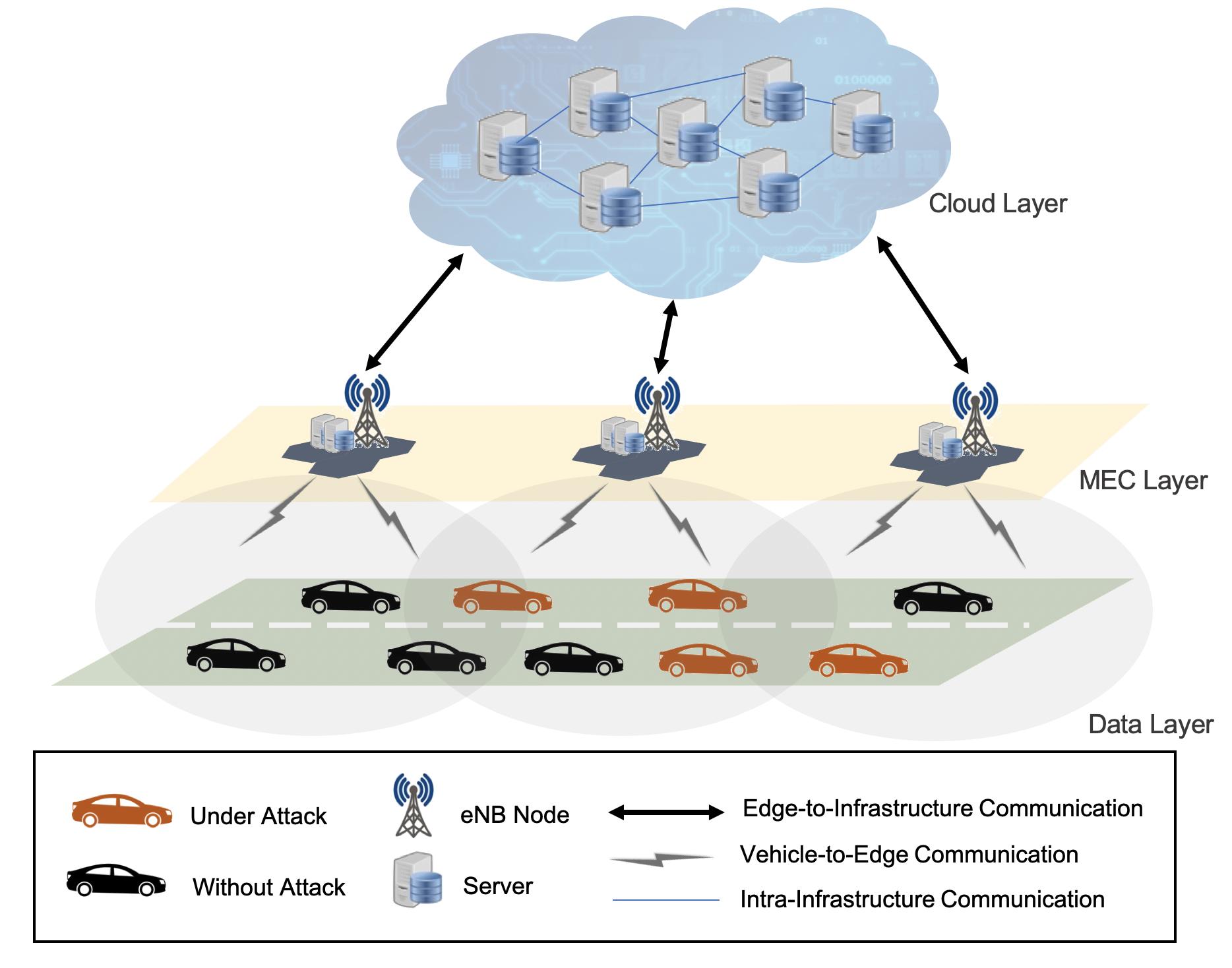}
		\caption{Three-layer IoV Network Architecture}
		\label{fig:architecture}
	\end{center}
\end{figure}
\vspace{-10pt}
DRL is a category of ML which combines deep learning (DL) and reinforcement learning (RL). RL is an objective-oriented algorithm. It learns a policy regarding how to achieve an objective and how to maximize it as system progresses. The actions which help to get the targeted objective are reinforced. In DRL, the policy is a neural network, and its framework contains an agent that interacts with the environment as it operates and makes a decision by rewarding the desired behavior and punishing the undesired one. The continual interaction of the DRL agent with the end-user exposes it to the number of adversarial attacks \cite{DRLattack,talpur2021machine}.  The assumption of having a secure environment to interact is not satisfactory in vehicular applications where misbehavior from an attacker can be life-threatening in case of accidents in vehicles. \par 
Data poisoning attacks like label flipping, backdoor attack, and model poisoning attack are very common adversarial attacks and explored in the literature for vehicular applications \cite{talpur2021machine}. In the context of DRL, where each time user feedback is collected to calculate reward, the effectiveness of an attack can increase using Sybils. A Sybil-based attack is an impersonation attack in which a malicious vehicle masquerades as a set of vehicles by stealing or borrowing the identities of legitimate users, or by creating fake ones \cite{sybilAttackPhases}. Sybil attack is easy to execute in mobile IoV networks where the communication is wireless, broadcast, and dynamic. When Sybil attack overloads the network with fake poisonous messages it can also lead to other attacks such as denial of service (DoS) attacks or replay attacks.  \par 
The Sybil defense schemes in the literature are not much effective for mobile networks due to the lack of historical behaviors in attack detection schemes \cite{SybilAttackDefensesIoT}. In addition, the traceability of Sybil nodes is also difficult because of the high mobility of vehicles. There are also proposals on using privacy-preserving schemes, trust-enabling, and other attack detection methods to protect against Sybil attacks \cite{sybildetection2,sybildetection3,sybildetection4}. Nevertheless, it is possible to craft well-trained intelligent Sybil attacks over the internet of things (IoT) which impose more serious threats and make it difficult to devise a defense mechanism to prevent or detect Sybil nodes \cite{sybilAttackPhases}. Not limited to this, there are studies which show the possibility of successful launching of Sybil attacks on mobile networks \cite{sybilattack}. Despite being a critical threat, the study on impacts of Sybil-based data poisoning attack and its defense against DRL-assisted mobile IoV network is not much explored.\par   
In this work, we study the problem of adversarial attacks on a DRL framework. Specifically, we study the Sybil-based data poisoning attacks in the context of DRL-based service placement mechanism in IoV, investigate the impacts on decision making and evaluate the delay perceived by vehicles and edge resource usage when the network is under attack. We consider a three-layer IoV architecture, as shown in Fig. \ref{fig:architecture}. The DRL framework is deployed at the Multi-access Edge Computing (MEC) layer. It consists of the DRL agent, where actor network and critic network are the primary functions to perform an action and evaluate decision quality value. The agent has direct interaction with the time-varying data layer of the IoV environment. In the proposed service placement technique, the agent receives delay as feedback from the environment to calculate quality values, and carry out decisions on effective service placement in a dynamic vehicular network. In case of data poisoning, the adversary tries to disrupt the network performance, misguides the DRL mechanism and affects it's decision by sending fake data (i.e. replacing long delays with small delays) in the feedback using Sybil nodes. The fake delays will lead to improper placement of services at edge nodes. To quantitatively analyze the impacts of wrong decision making on service delay and edge resource usage, we carry out a comprehensive experimental study with real-world vehicle trajectories. We further investigate the impact of the proportion of Sybil-attacked vehicles with different choices of types of services under attack when compared to attack-free healthy network scenario.

\section{System Model}
\label{Sec:model}
The hierarchical architecture of our proposed IoV network model is shown in Fig. \ref{fig:architecture}. It consists of three layers that include data layer, MEC layer, and cloud layer. At the data layer, a city road environment with a real journey of taxis in San Francisco is used. The vehicles are mobile, which are assumed to generate service requests with a certain rate for a type of service. A service is a facility like CAM (cooperative awareness message) service, teleoperated support service, media downloading/sharing service, etc. Vehicles use these services to get assistance in different tasks related to driving. Each service/application has its own deployment requirements in terms of resource and delay defined by the standardizing bodies for well-timed and effective provisioning of a facility. \par 
We also assume each vehicle $\nu$ is equipped with necessary sensors like clock and GPS, which enable it to provide relevant information. The attacks are assumed to take place at the data layer. The attack model is discussed in Section \ref{Sec:attackmodel}. For the MEC layer, we assume the IoV network environment is under 5G coverage using evolved NodeB (eNB) stations. There are multiple eNBs equipped with edge servers that extend the capabilities (storage and compute) of the cloud and bring them closer to the end user. The resources available at the edge required to provide services are limited. A DRL-based dynamic service placement algorithm (discussed in Section \ref{Sec:DRL}) is deployed at the MEC layer to facilitate vehicles with multiple requested services. Additionally, the edge network connects to the large capacity cloud layer via a backbone network. We assume adequate links between different layers, nodes, and servers are available to enable communication among them.

\section{DRL-Assisted Service Placement Framework}
\label{Sec:DRL}
\begin{figure}[htbp]
	\begin{center}
		\includegraphics[width=3.1in,height=1.5in]{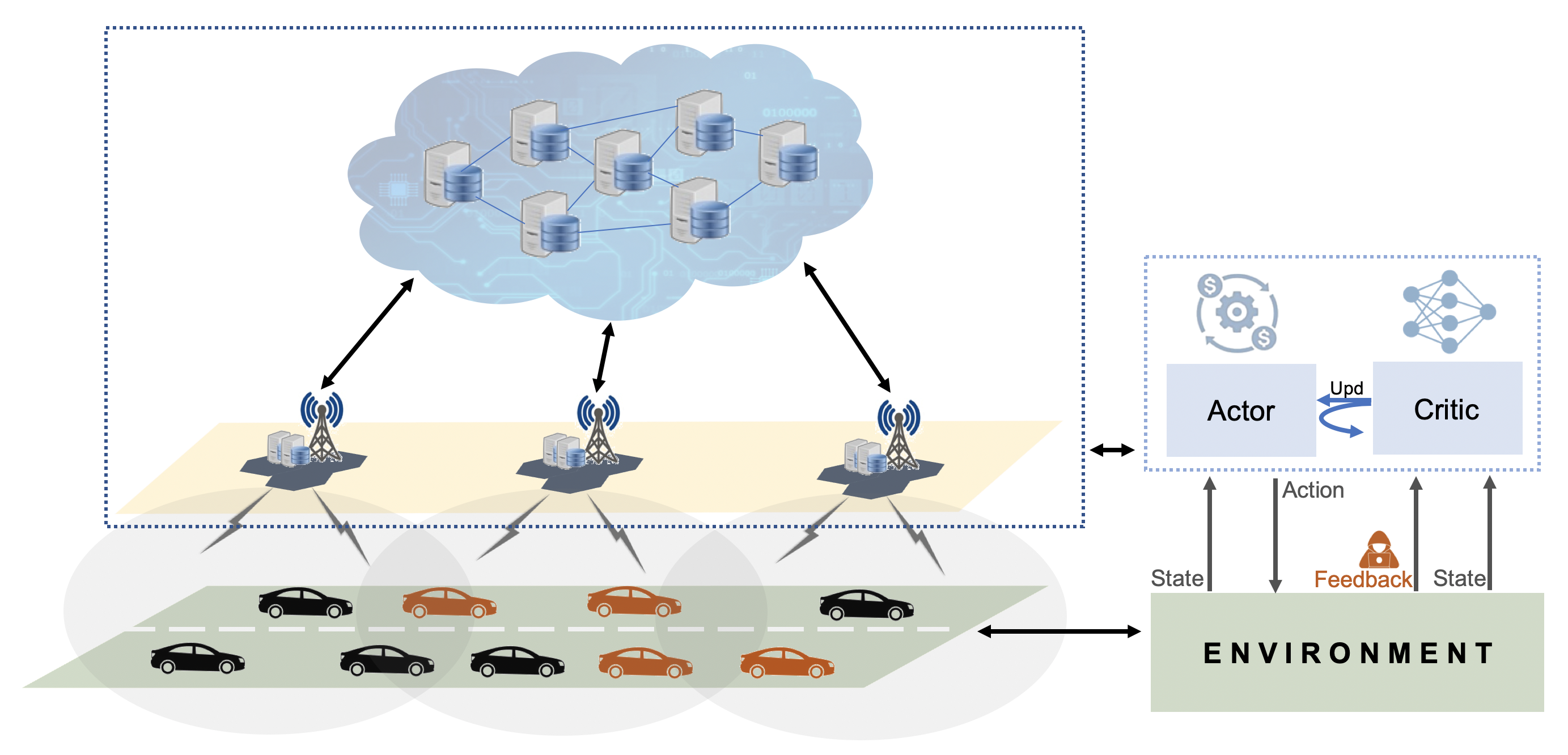}
		\caption{DRL-assisted Service Placement Framework}
		\label{fig:model}
	\end{center}
\end{figure}

In our recent work \cite{talpur2021drld}, we developed a DRL-based dynamic service placement framework for IoVs. In this paper, we use this framework to study the problem of Sybil-based data poisoning attacks and evaluate the impact on vehicular services. The architecture of the service placement approach is shown in Fig. \ref{fig:model}. \par 
\vspace{-10pt}
The framework consists of a DRL agent over the edge network. Here, the \textbf{actor network} and \textbf{critic network} are the agent's primary functions to perform an action and evaluate decision quality value. The agent has direct interaction with the time-varying IoV environment. We use an ILP-based optimization at the \textbf{actor network} to perform optimal service placement. Further, the design involves \textit{state space $\omega$, action space $\mathfrak{a}$, policy, and reward/feedback function $\mathcal{R}$}, as shown in Fig. \ref{fig:model}. The \textit{state space $\omega$} describes the network environment in terms of the service request message. A service request is defined as a 4-tuple structure containing the information on vehicle ID, type of service requested, location of the vehicle, and time. In return, considering the demand at a given time and the location of vehicles requesting services, the policy function comes into action to select the edge servers for the placement of services based on the action strategy. Therefore, the \textit{action space $\mathfrak{a}$} describes the action taken by a policy function for the placement of service over an edge node.\par 
In this framework, the \textit{policy function} optimizes the objective function subject to different constraints. The objective is to minimize the maximum edge resource usage and service delay, with the control of the relative importance of resource usage vs. service delay by using a parameter $\alpha$. The rationale for using resource usage is to efficiently utilize the limited edge resources and decrease the possibility of congestion so that the edge node has enough room for service instance scale-up in case of increased future demands. From the perspective of a user, minimizing the maximum delay will help to satisfy adequate delay requirements and make service availability faster for the vehicles. At each time unit, in response to the action taken by an actor network, the system receives an immediate \textit{reward} $\mathcal{R}(\omega,\mathfrak{a})$ from the environment. The reward/feedback function used in our framework is calculated as:
\begin{equation}
	\mathcal{R}(\omega,\mathfrak{a})=\mathbb{E}\left[d_e^s(t)\right]
	\label{eq:reward}
\end{equation}
where $d_e^s(t)$ is the average service delay observed by a set of vehicles in accessing service $s$ from the associated edge server $e$ at time unit $t$. The objective of reward values in our framework is to maintain low service delay observed from vehicles in accessing a service. Here, the reward is an important parameter to enhance the performance of the time-varying IoV network for future actions. This impacts the performance of the DRL framework in the case of false information due to attacks over the environment \par 

Further, the \textbf{critic network} is responsible for calculating the quality value $Q(\omega,\mathfrak{a})$ of the decision taken by the actor network. A high $Q(\omega,\mathfrak{a})$ means a high-quality decision. Therefore, an actor has to select actions with the maximum quality value, $\mathfrak{a}=$ arg max $Q(\omega, \mathfrak{a})$. However, the performance of the critic network highly depends on the feedback (rewards) from the environment. In this design, the critic network is a neural network, and the input of the neural network include a state $\omega$, action $\mathfrak{a}$, and reward $\mathcal{R}$. The critic network updates its parameters $\theta$ to minimize the mean square loss function $\mathcal{L}_Q$. The loss function is computed as:
\begin{equation}
	\mathcal{L}_Q(\theta)=\frac{1}{\mathcal{N}}\sum_{i=1}^{\mathcal{N}}\left[(y_{t_i}-Q_i(\omega,\mathfrak{a};\theta))^2\right]
	\label{eq:loss}
\end{equation}
Here, $y_t$ is a target value and $\mathcal{N}$ is the batch size used to update the critic network parameters. The DRL agent further uses a replay memory. It is used to store the experience for training the critic network. The critic network uses replay memory to fetch experience after a random period $T$ and optimizes the network parameters for better performance. Once the network is trained, the procedure for decision making gets simple. Altogether, the trained critic network is used by the DRL agent to observe the state and perform an action for which the quality value is maximum. Later, it obtains a reward and observes a new state to facilitate traffic for the next time unit and so on.

\section{Sybil-based Data Poisoning Attack Model}
\label{Sec:attackmodel}
The data poisoning attack is an integrity attack which pollutes the ML model's input data and impacts its ability of decision making. In the case of DRL-assisted service placement, the vehicle to edge (V2E) communication is responsible for providing feedback to the DRL. The data poisoning attack over V2E communication is a serious threat where the decisions are based on real-time interaction with the environment and the effectiveness of an attack can increase using Sybils. Here, the Sybil node is an impersonation node in which a malicious vehicle masquerades as a set of vehicles by stealing or borrowing the identities of legitimate users. It is hard to detect and easy to execute in mobile IoV networks \cite{SybilAttackDefensesIoT}. The attacker can easily degrade the network performance with data poisoning and disrupts the services which are crucial for driving-related decisions. \par 
There can be trained attackers or amateur attacks. Sybil nodes are also classified into three different classes, i.e. SA-1, SA-2, and SA-3 \cite{SybilAttackDefensesIoT}, depending on the type of network under attack. Here, SA-1 and SA-2 are for static networks. The SA-3 type is for mobile networks. Due to the dynamic nature of the IoV environment, in this work, we use SA-3 where Sybil nodes are mobile and distributed throughout the network. We assume a compromised-node can launch SA-3 attacks where the identities of legitimate vehicles are purposely stolen. The stolen identities can easily be hidden and remain undetected if the imitated vehicles are temporarily restricted from the network. We assume that the adversary can gain access to all the information related to a vehicles whose IDs are stolen \cite{sybilAttackPhases}. Moreover, compromised identities will also help an attacker to easily pass all security checks of the network. The adversary is then able to reprogram the vehicle to behave maliciously by performing data poisoning and sending fake information to the network. In the case of service placement, we use fake reward values (i.e. replace long delays with small delays) which are sent back to the DRL agent to make it taking wrong decisions on quality values, which will lead to improper placement of services at the edge.\par
\begin{figure}[htbp]
	\begin{center}
		\includegraphics[width=3in,height=2.2in]{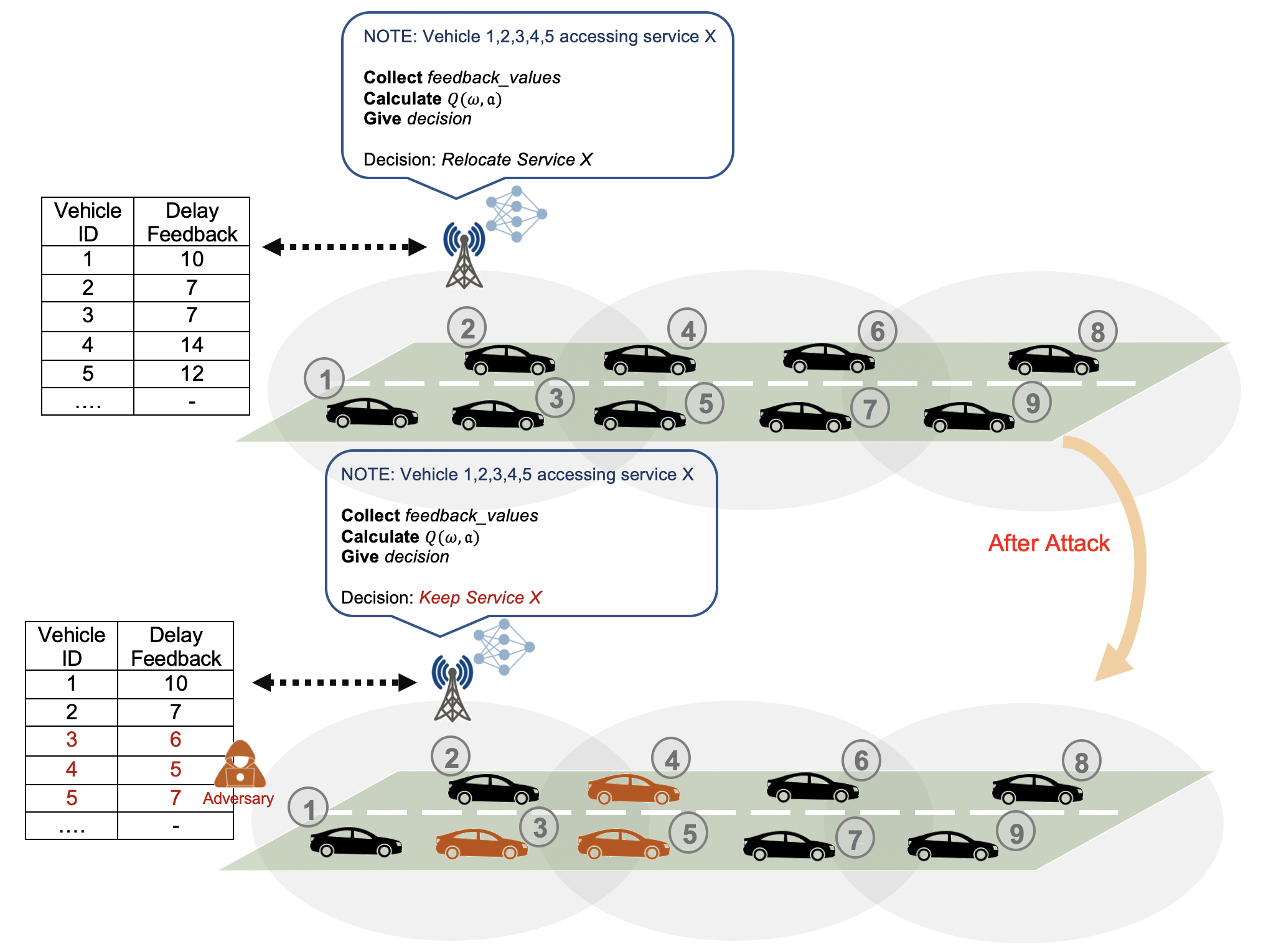}
		\caption{Illustration of the Sybil attack in IoV}
		\label{fig:example}
	\end{center}
\end{figure}

 
Fig. \ref{fig:example} illustrates an example scenario of a Sybil-based data poisoning attack over three vehicles. The number inside the small circle represents the vehicle's identity. We assume the vehicles with identity numbers 3, 4, and 5 are under a Sybil attack. We further suppose that vehicles 1, 2, 3, 4, and 5 are under coverage and accessing \textit{service X} from the edge node. When there is no attack, the table shows a high delay observed by three vehicles out of five vehicles. Considering this, it can be observed by the edge node that many vehicles are leaving the coverage which requires relocation of service X or installation of a new instance of service X at another edge node, which is closer to the vehicles. The new location for service X will be calculated by the actor network, as discussed in Section \ref{Sec:DRL}. On the contrary, when the network is under attack, the attacker intrudes the network by compromising vehicles 3, 4 and 5, and uses these identities to launch data poisoning and send fake delay values in the feedback. The fake values are smaller delay values from the valid range of possible observed delay values in the given scenario. This obscures the edge network, and it interprets that the network performance is good and vehicles can continue service X from the same edge node. Thus the vehicles continue to receive poor service which is undesirable. 
\vspace{-16pt}

\section{Performance Evaluation}
\label{Sec:simmulationstudies}
We carry out a comprehensive performance study on the real-world dataset to evaluate the impacts of attack. 
\subsection{Experimental Setup}
\begin{figure}[htbp]
	\begin{center}
		\includegraphics[width=2.1in,height=1.5in]{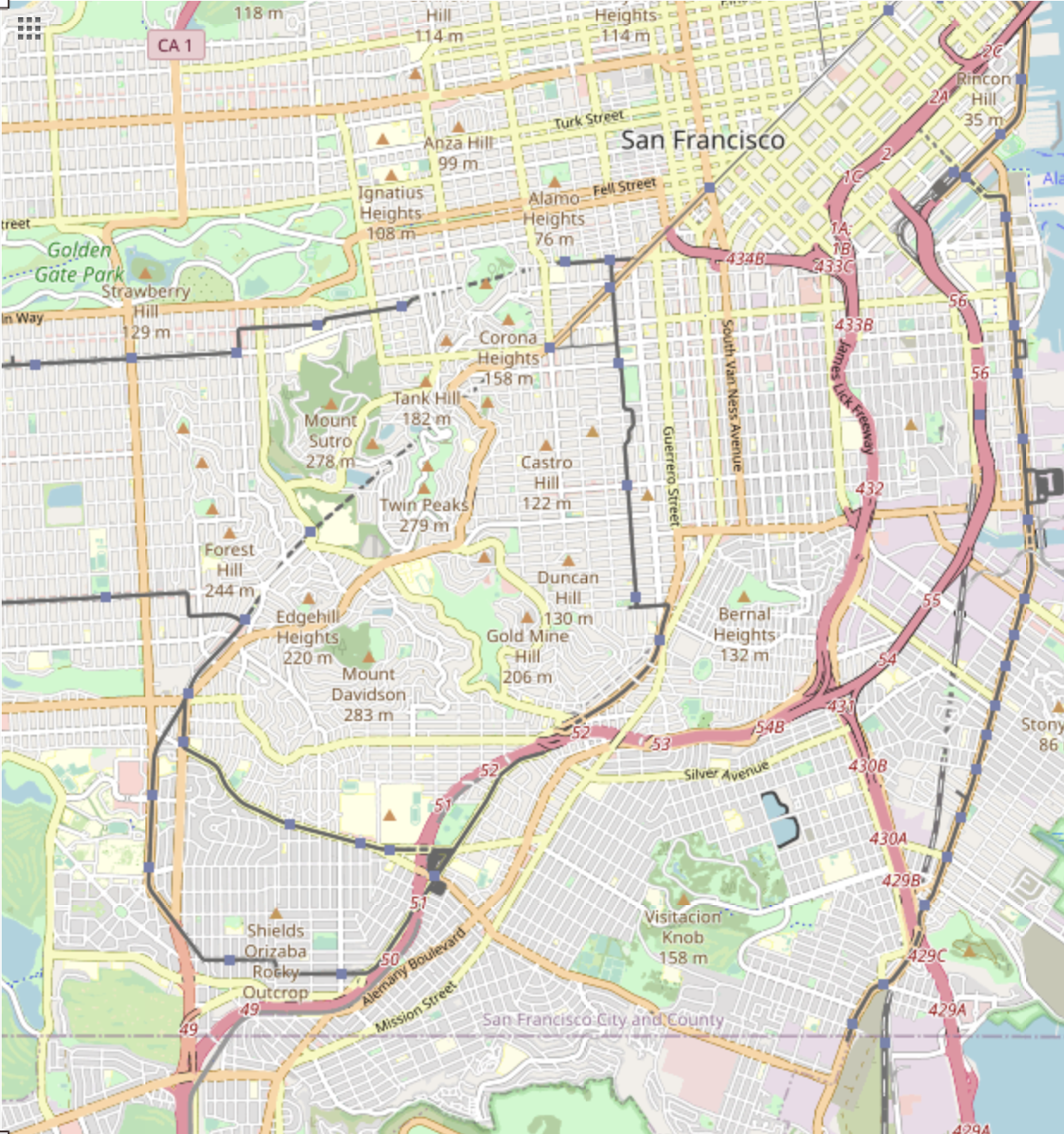}
		\caption{The experimental coverage area}
		\label{fig:sanfrancisco}
	\end{center}
\end{figure}

The simulations are carried out using the MATLAB platform. The vehicle trajectories are obtained from a real-world vehicle mobility dataset, provided by crawdad \cite{dataset}. The data is generated from 500 taxis traveling the city of San Francisco. The choice of the dataset is significant, as it is an urban environment with high traffic densities. From the big city area given, we extract an area of $10x10{km}^2$ for use in our experiments, as shown in Fig. \ref{fig:sanfrancisco}. Each taxi is equipped with a GPS sensor and uploads its geo-coordinates record in real-time to form the vehicle trajectories. Each location updated to the central server includes a timestamp, vehicle identifier, and geo-coordinates. \par 
Further, a uniformly distributed and randomly chosen choice of service is added to a record to form a service request message sent from each vehicle to the edge network. In total, there are 8 types of services with their pre-defined resource requirements $R_s$ and delay thresholds $D_s$, as shown in Table. \ref{tab:sim-parameters}. The delay threshold represents the maximum allowable delay for each service. The fake delay value $f_d$ is chosen from the range of values used to poison original values. The DRL agent which receives the service request message is deployed at the edge. At the MEC layer, there are 6 eNBs, each equipped with edge servers with the capability to place multiple services while satisfying the resource requirements, i.e., the total resources consumed by services $\sum R_s$ to be placed must be less than the available resources at that particular eNB server $C_e$. The design parameters for DRL agent neural network design and optimization algorithm are the same as discussed in \cite{talpur2021drld}. \par
\vspace{-9pt}
In our experiment, first, the attack-free network scenario is considered wherein all vehicles are assumed to be legitimate and the network performs effective service placement decisions. The results collected in attack-free scenarios are called as \textbf{\textit{no-attack (NA)}} in this paper. In case of an attack, different proportions of vehicles \textbf{\textit{under-attack (UA)}} are chosen for performing multiple experiments. At first, the Sybil attacker acquires a set of valid identities, called Sybil identities, in the \textit{node compromise phase}. The next phase is the attack \textit{deployment phase}, wherein we randomly pick locations to deploy malicious nodes throughout the network. We do not restrict all attacked vehicles within a limited region or the same neighborhood. Finally, to \textit{launch attack}, the fake delay (but in the valid range) information (i.e. low delay values) is sent back to the DRL agent as a reward for Sybil nodes. We assume the attacker is smart enough to calculate the valid range of delay in a given scenario.\par 
\begin{table}[htbp]
	\centering
	\caption{Simulation Parameters}
	\begin{tabular}{|l|l|l|l|}
		\hline
		\textbf{Parameters} & \textbf{Value} & \textbf{Parameters} & \textbf{Value} \\
		\hline
		Services     & 8 & $R_s (Unit)$ & [5 10 15 20 25 30 35 40] \\
		Vehicles     & 500 & $C_e (Unit)$ & [60 70 80 90 100 100] \\
		eNBs    & 6 & $D_s (ms)$ & [14 16 18 20 22 24 26 28] \\
		time  & 1 to 900 & $f_d (ms)$ & \{3,4,5,6,7,8,9\} \\
		\hline
	\end{tabular}%
	\label{tab:sim-parameters}%
\end{table}% 
\vspace{-9pt}

\subsection{Results and Discussion}
In this section, we evaluate the impacts of the attack in terms of five different performance metrics that include, number of re-optimizations, average service delay, average service delay of targeted vehicles, edge resource usage, and fairness (Jain's index \cite{jain_index}). We investigate the effect for different proportion of Sybil vehicles. We use 10\%, 20\%, 30\%, 40\%, and 50\% as the proportion of vehicles under attack. Further, we study two scenarios; \textbf{\textit{case 01}} is "\textit{\textbf{Attack-Any}}", we consider all vehicles with any type of service request are under attack; \textbf{\textit{case 02}} is "\textbf{\textit{Attack-Selective}}", we assume selected vehicles, i.e. vehicles requesting for service 1 to 4 only, are under attack. For each experiment, we conduct 5 runs with different random seeds and plot the average results.
\par
\begin{figure}[hbt!]
	\centering
	\begin{subfigure}{.23\textwidth}
		\centering
		\includegraphics[width=1.7in]{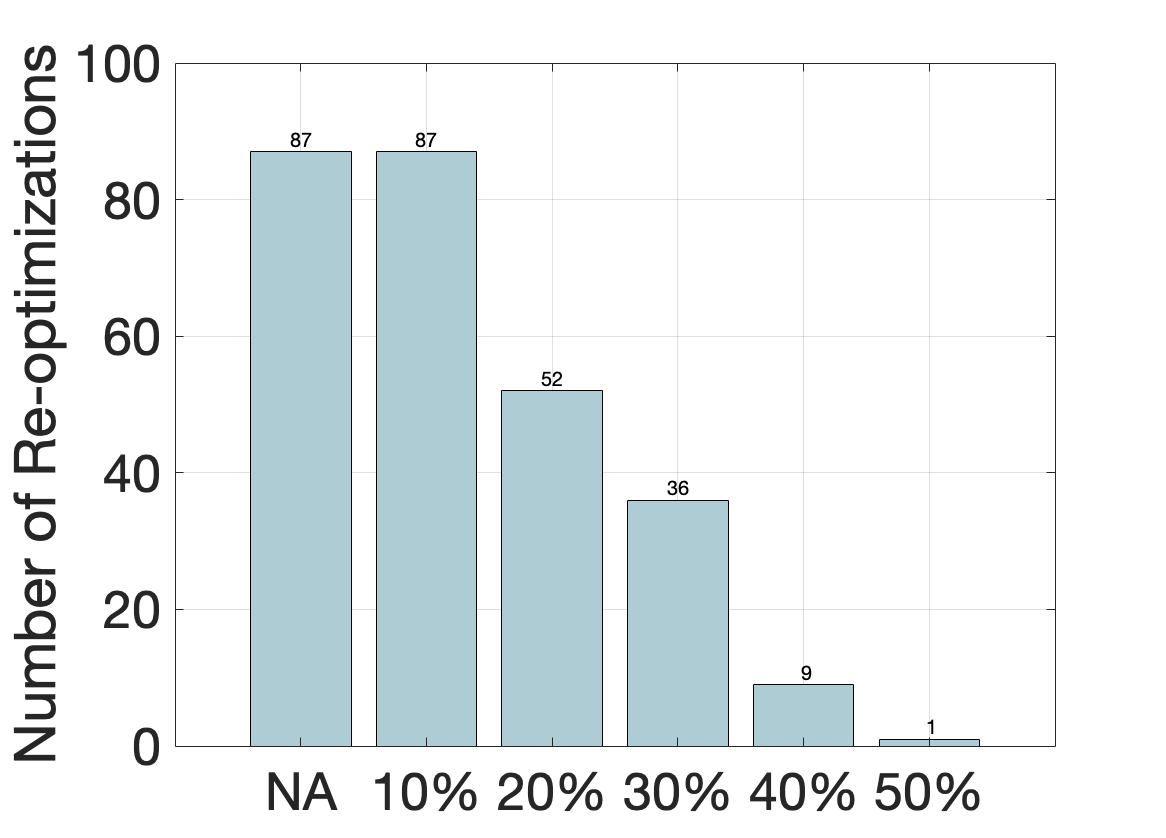}  
		\caption{Attack-Any}
		\label{fig:reopt1}
	\end{subfigure}
	\begin{subfigure}{.23\textwidth}
		\centering
		\includegraphics[width=1.7in]{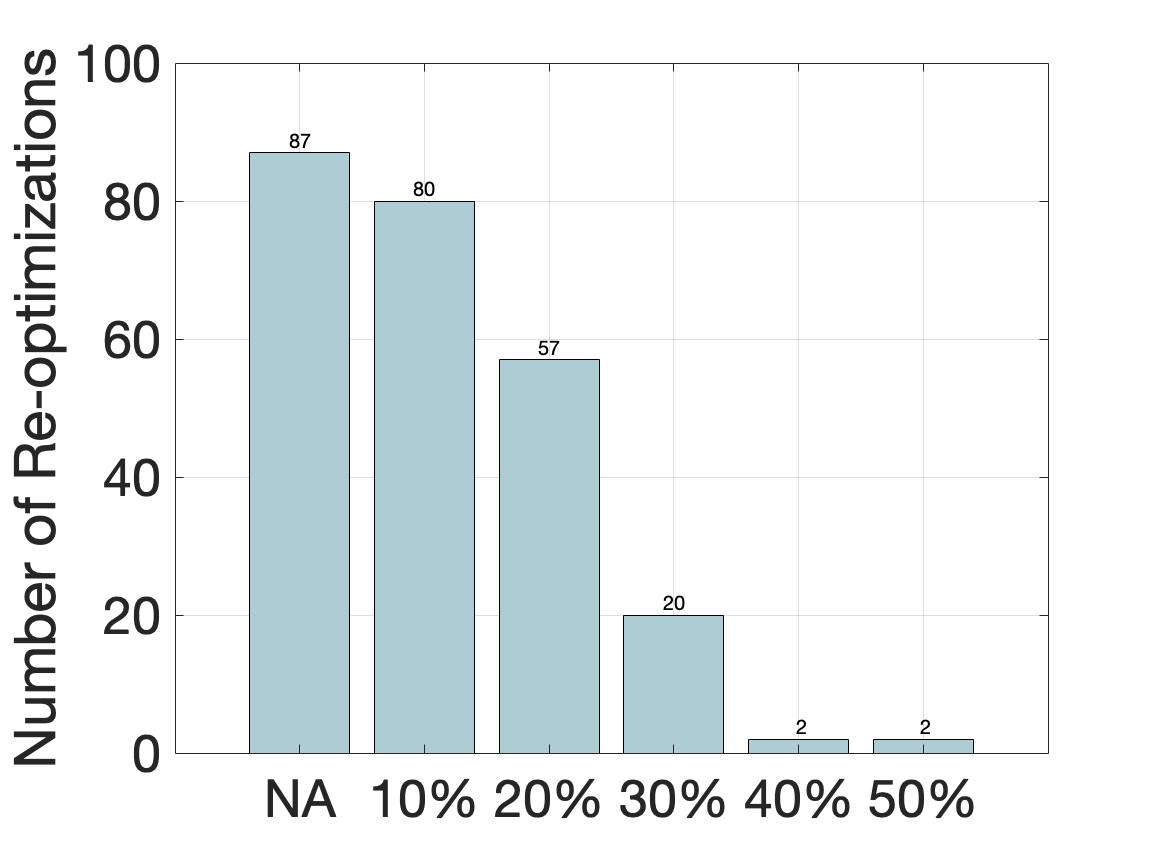}  
		\caption{Attack-Selective}
		\label{fig:reopt2}
	\end{subfigure} 
	\caption{Number of Re-optimizations}
	\label{fig:reopt}
\end{figure}

 
We first investigate the impact of different proportions of attacked vehicles on the number of re-optimizations in both cases, as shown in Fig. \ref{fig:reopt}. This metric gives the number of times the framework re-optimizes within the total duration of the experiment. The first bar with the label of no-attack (NA) scenario shows that in the absence of attacks, an average of 87 re-optimizations are needed to maintain good performance in both attack-any and attack-selective cases. On the other hand, in case of an attack, with the increasing proportion of attacked vehicles the number of re-optimizations decreases. This is because data poisoning with low delay values make the DRL think that high level of service quality is perceived by the vehicles. This force the critic network to give a high-quality decision, which makes the agent to decide on continuing with the same service locations for most of the time. When the 50\% of traffic is under attack, the number of re-optimizations reduces to 1 and 2 in attack-any and attack-selective case, respectively. This renders the DRL framework's functioning completely ineffective. \par
\vspace{-9pt} 
\begin{figure}[hbt!]
	\centering
	\begin{subfigure}{.23\textwidth}
		\centering
		\includegraphics[width=1.8in]{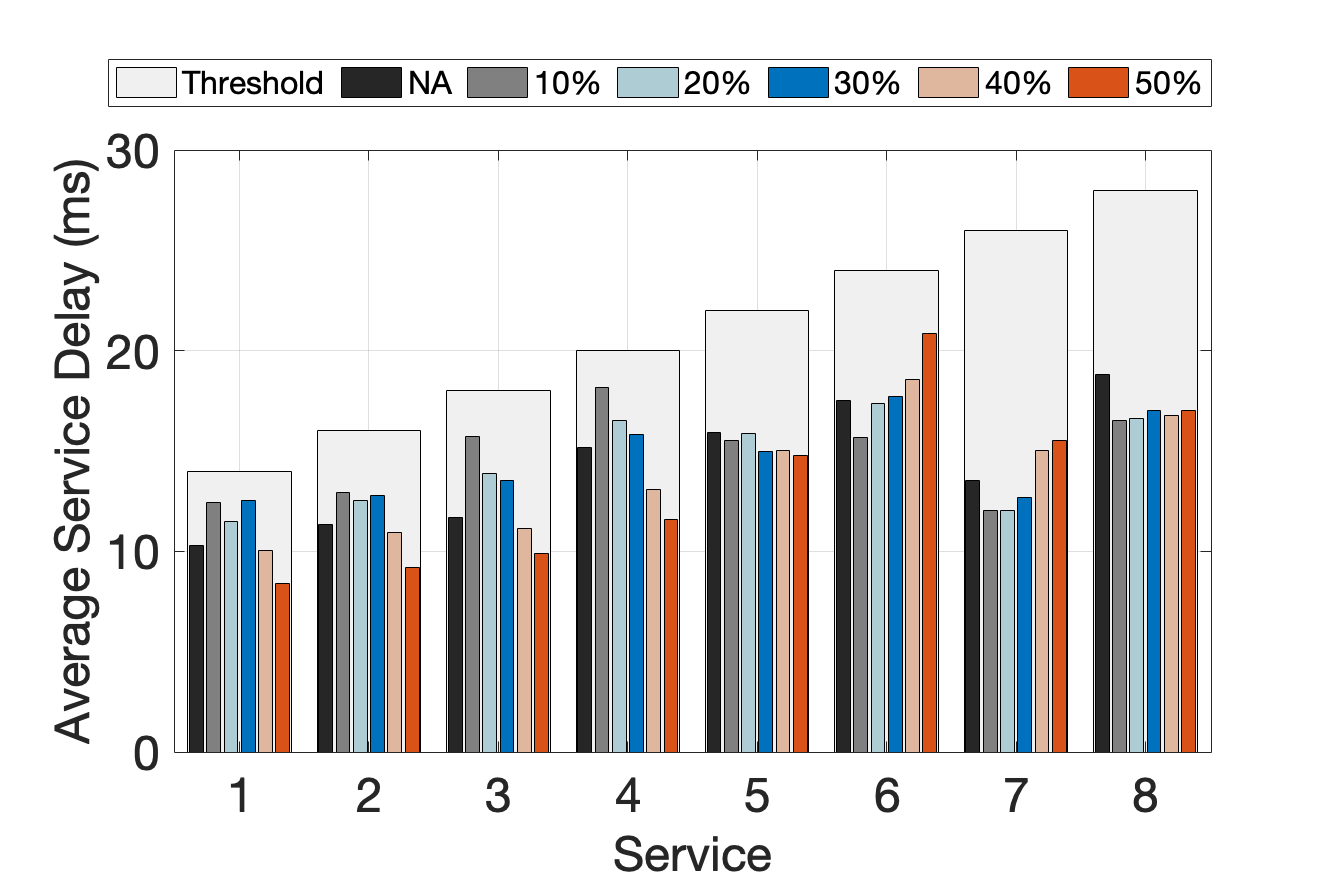}  
		\caption{Attack-Any}
		\label{fig:Avgdelay1}
	\end{subfigure}
	\begin{subfigure}{.23\textwidth}
		\centering
		\includegraphics[width=1.8in]{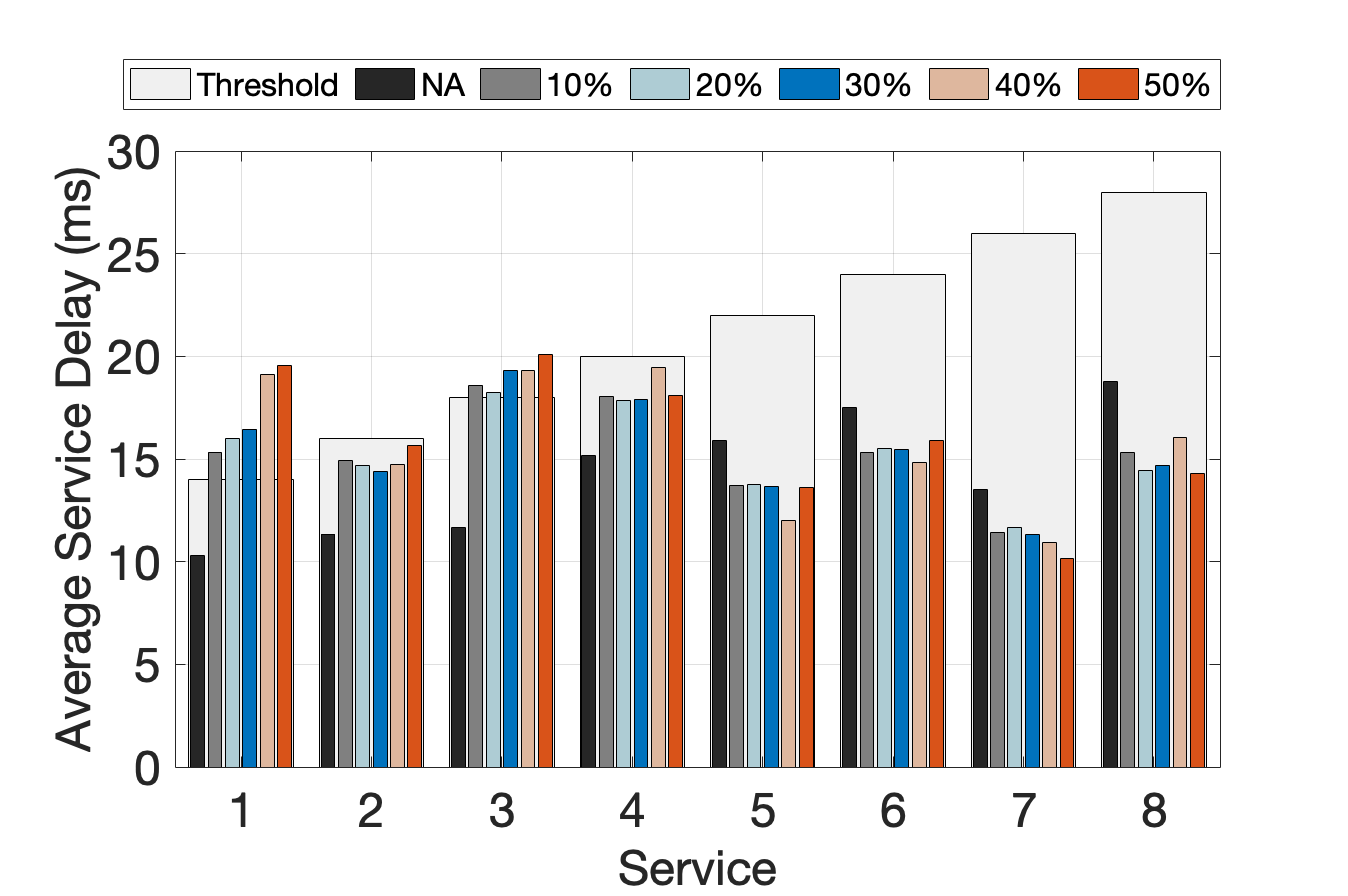}  
		\caption{Attack-Selective}
		\label{fig:Avgdelay2}
	\end{subfigure} 
	\caption{Average Service Delay}
	\label{fig:Avgdelay}
\end{figure}

\begin{figure*}[hbt!]
	\centering
	\begin{subfigure}{.19\textwidth}
		\centering
		\includegraphics[width=1.5in]{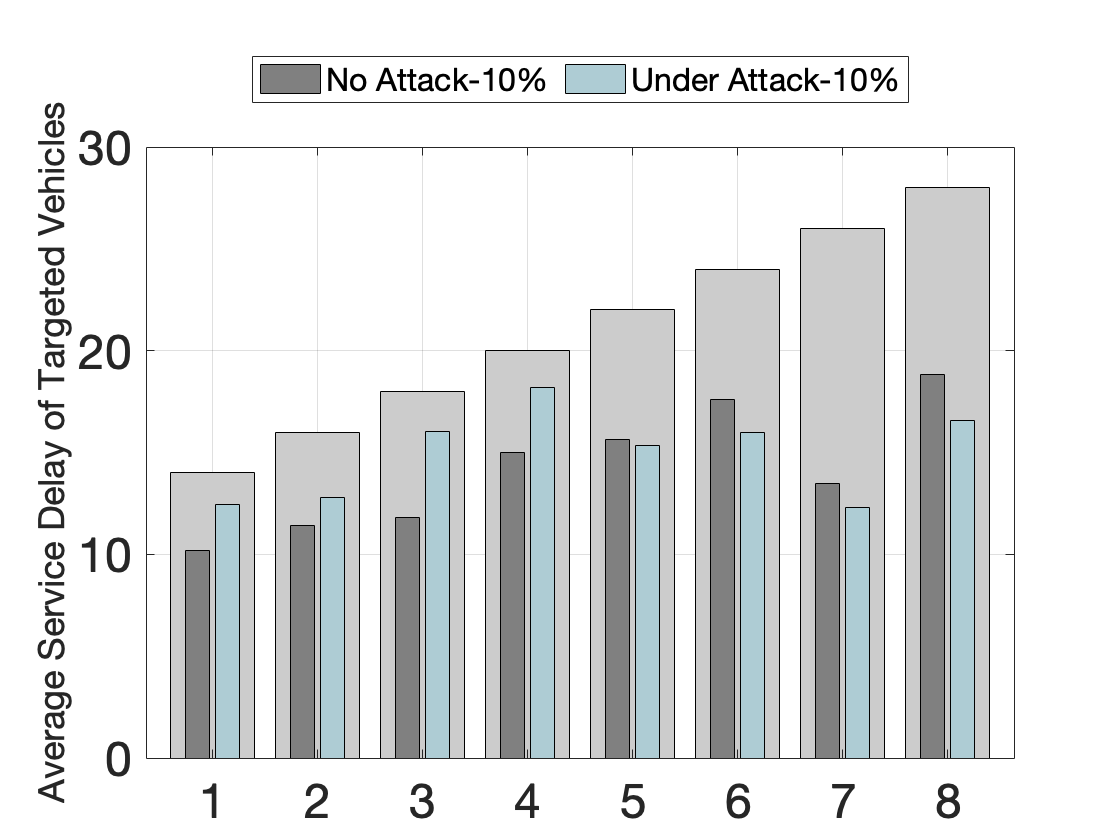}  
		\caption{ }
		\label{fig:1Attdelay10}
	\end{subfigure}
	\begin{subfigure}{.19\textwidth}
		\centering
		\includegraphics[width=1.5in]{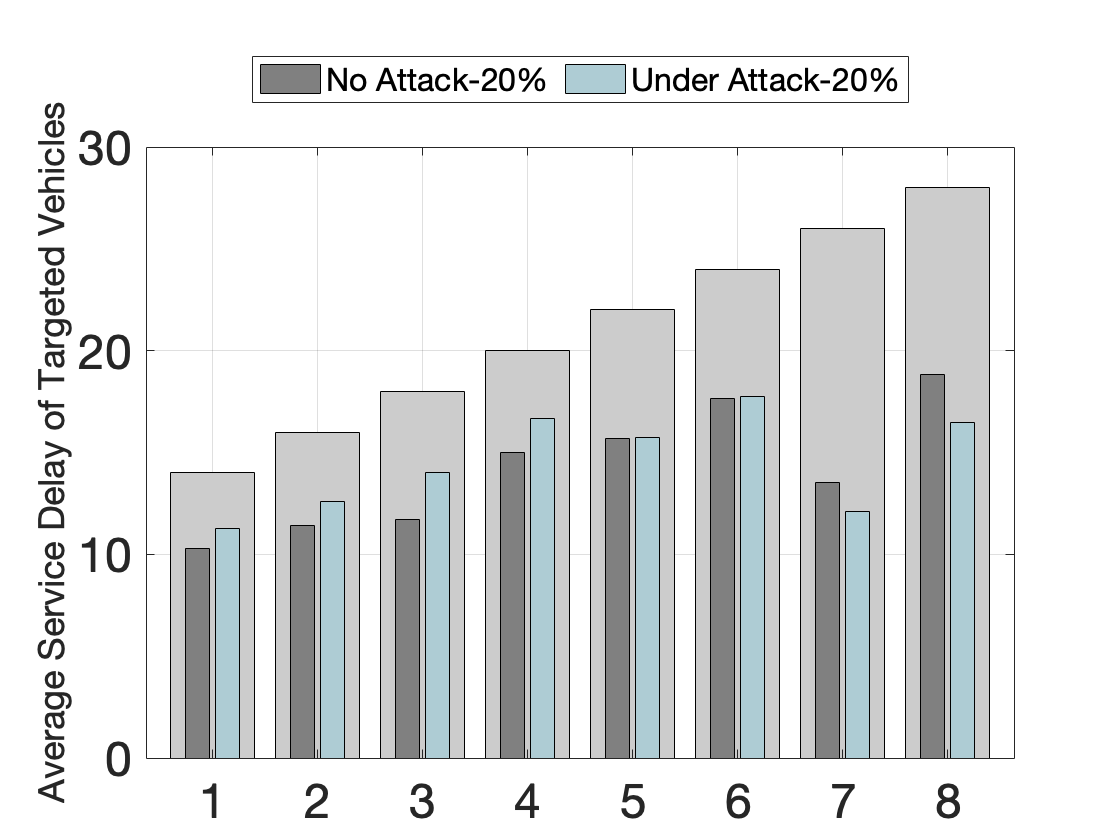}  
		\caption{ }
		\label{fig:1Attdelay20}
	\end{subfigure} 
	\begin{subfigure}{.19\textwidth}
		\centering
		\includegraphics[width=1.5in]{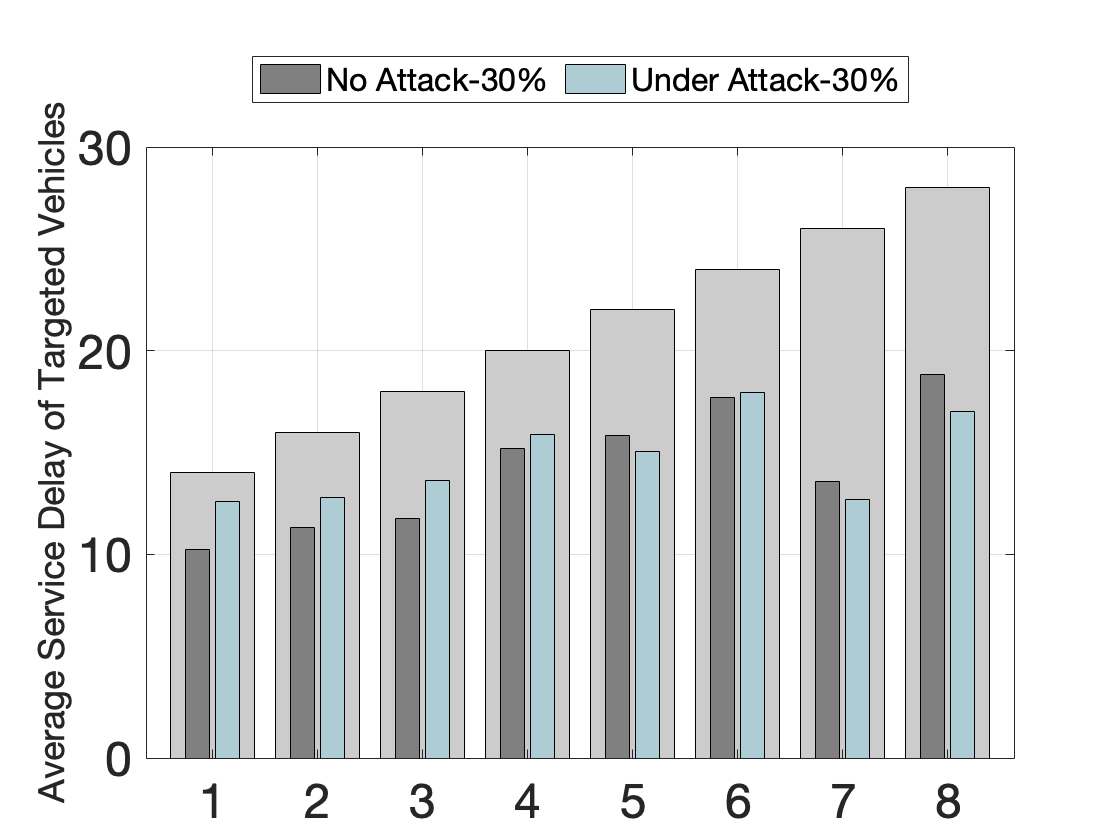}  
		\caption{ }
		\label{fig:1Attdelay30}
	\end{subfigure} 
	\begin{subfigure}{.19\textwidth}
		\centering
		\includegraphics[width=1.5in]{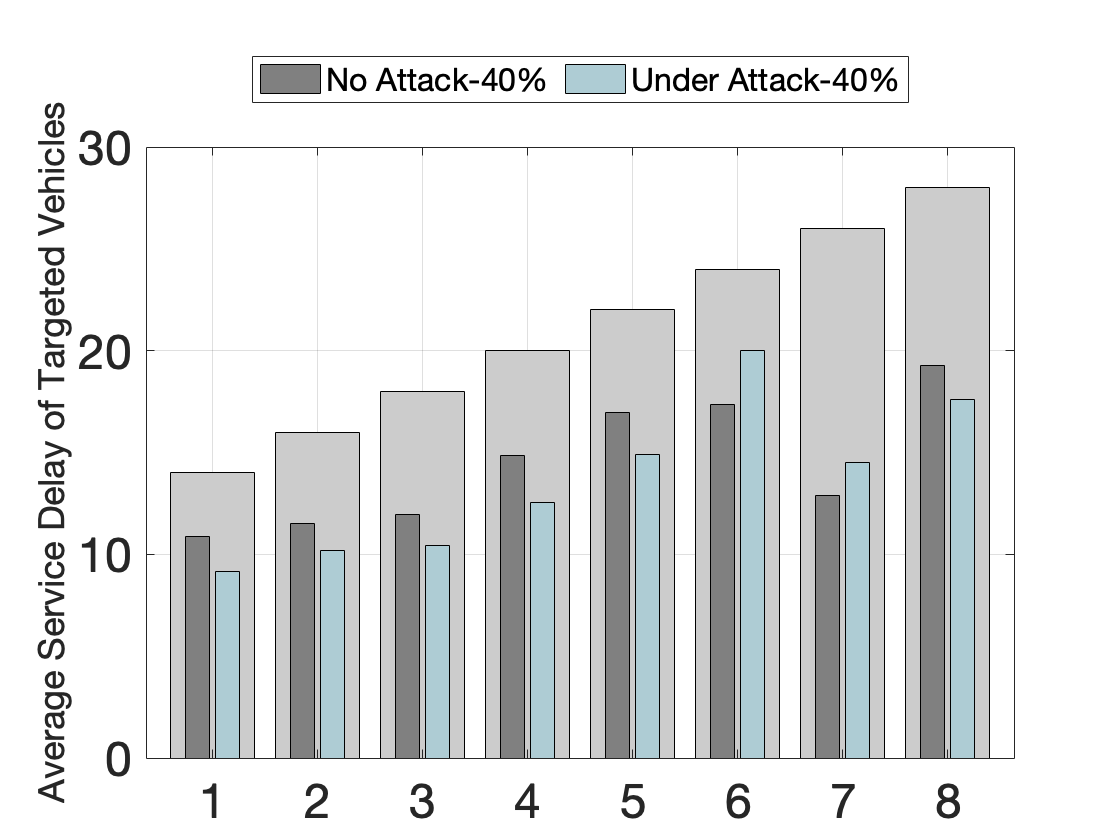}  
		\caption{ }
		\label{fig:1Attdelay40}
	\end{subfigure} 
	\begin{subfigure}{.19\textwidth}
		\centering
		\includegraphics[width=1.5in]{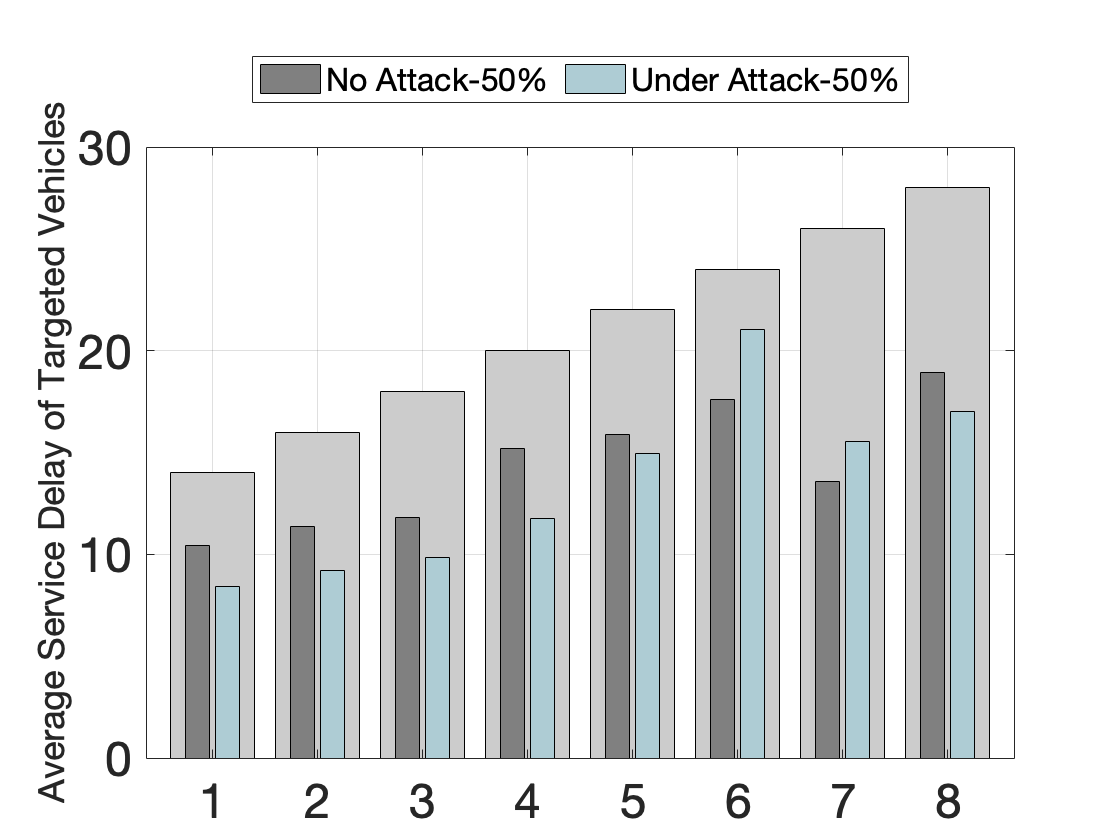}  
		\caption{ }
		\label{fig:1Attdelay50}
	\end{subfigure} 
	\caption{Average Service Delay of Targeted Vehicles - Attack-Any}
	\label{fig:Attdelay01}
\end{figure*}

\begin{figure*}[hbt!]
	\centering
	\begin{subfigure}{.19\textwidth}
		\centering
		\includegraphics[width=1.5in]{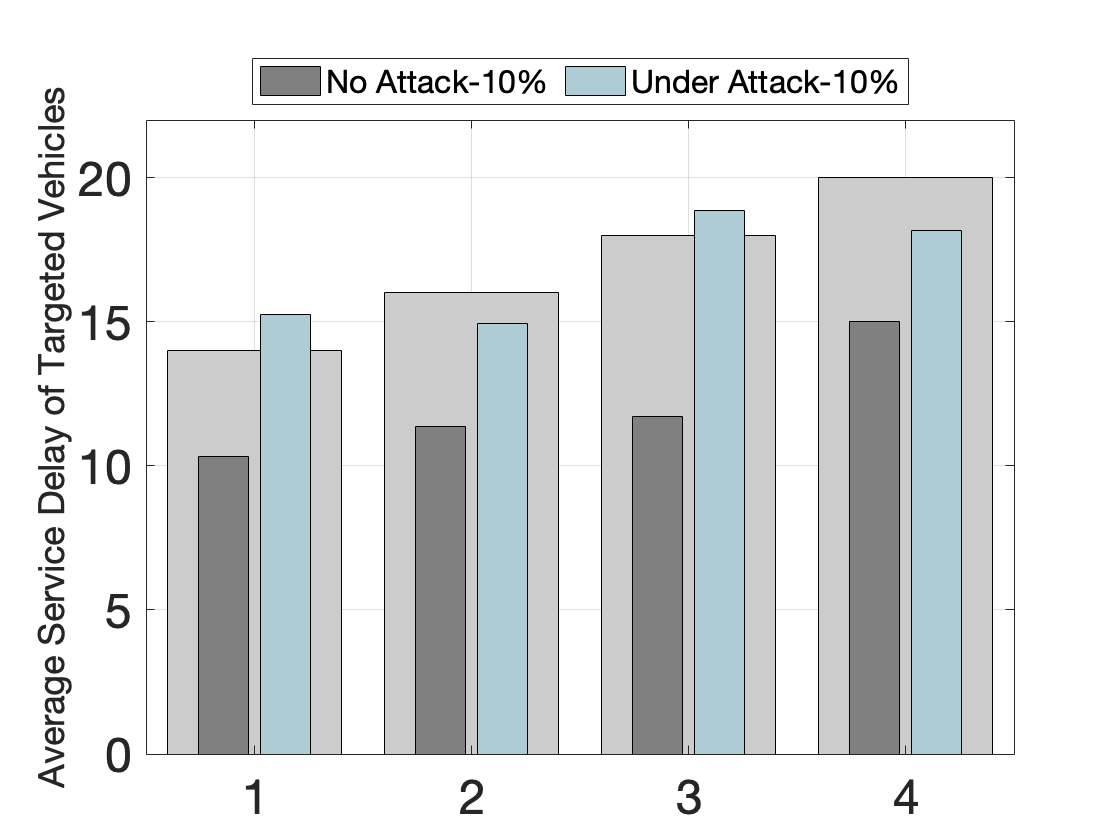}  
		\caption{ }
		\label{fig:2Attdelay10}
	\end{subfigure}
	\begin{subfigure}{.19\textwidth}
		\centering
		\includegraphics[width=1.5in]{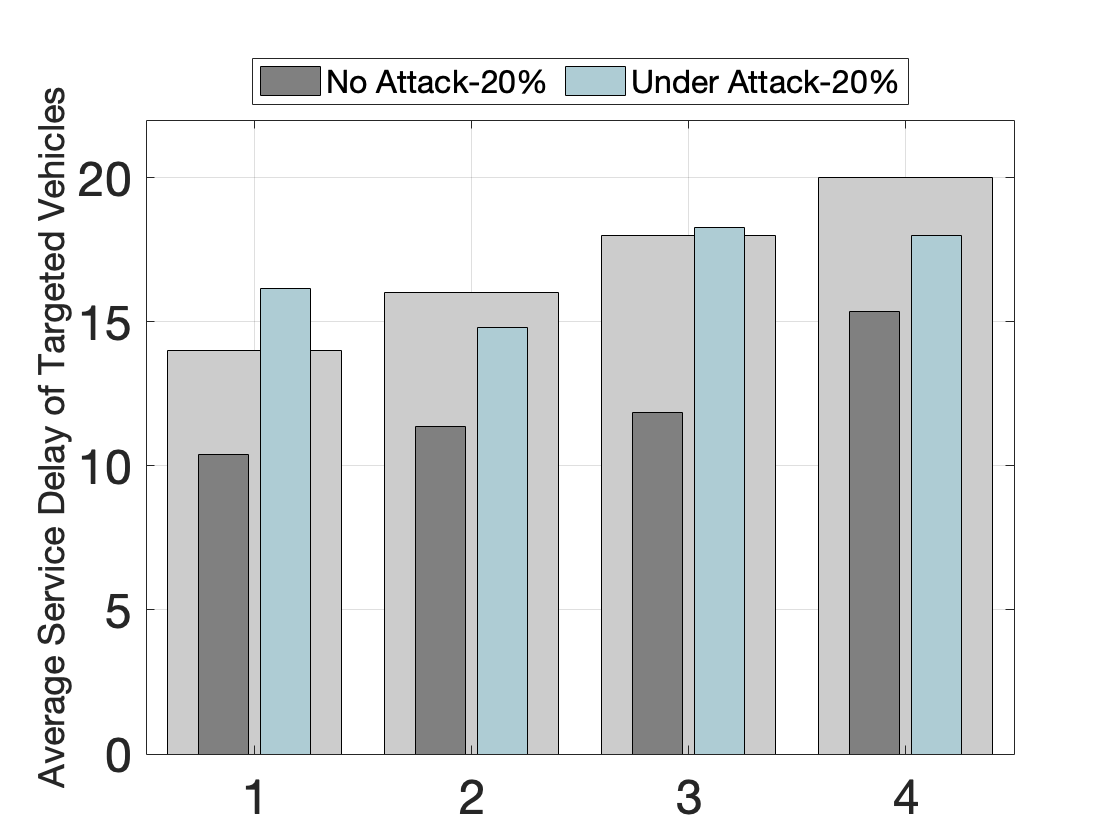}  
		\caption{ }
		\label{fig:2Attdelay20}
	\end{subfigure} 
	\begin{subfigure}{.19\textwidth}
		\centering
		\includegraphics[width=1.5in]{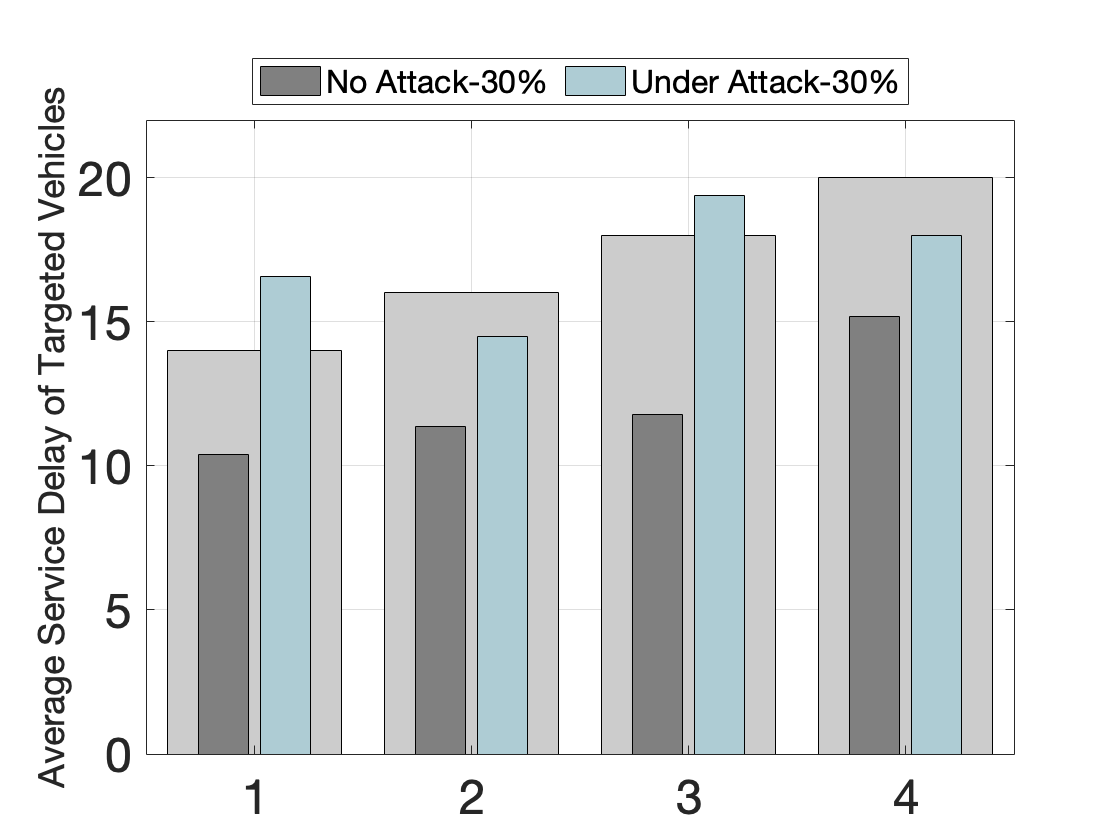}  
		\caption{ }
		\label{fig:2Attdelay30}
	\end{subfigure} 
	\begin{subfigure}{.19\textwidth}
		\centering
		\includegraphics[width=1.5in]{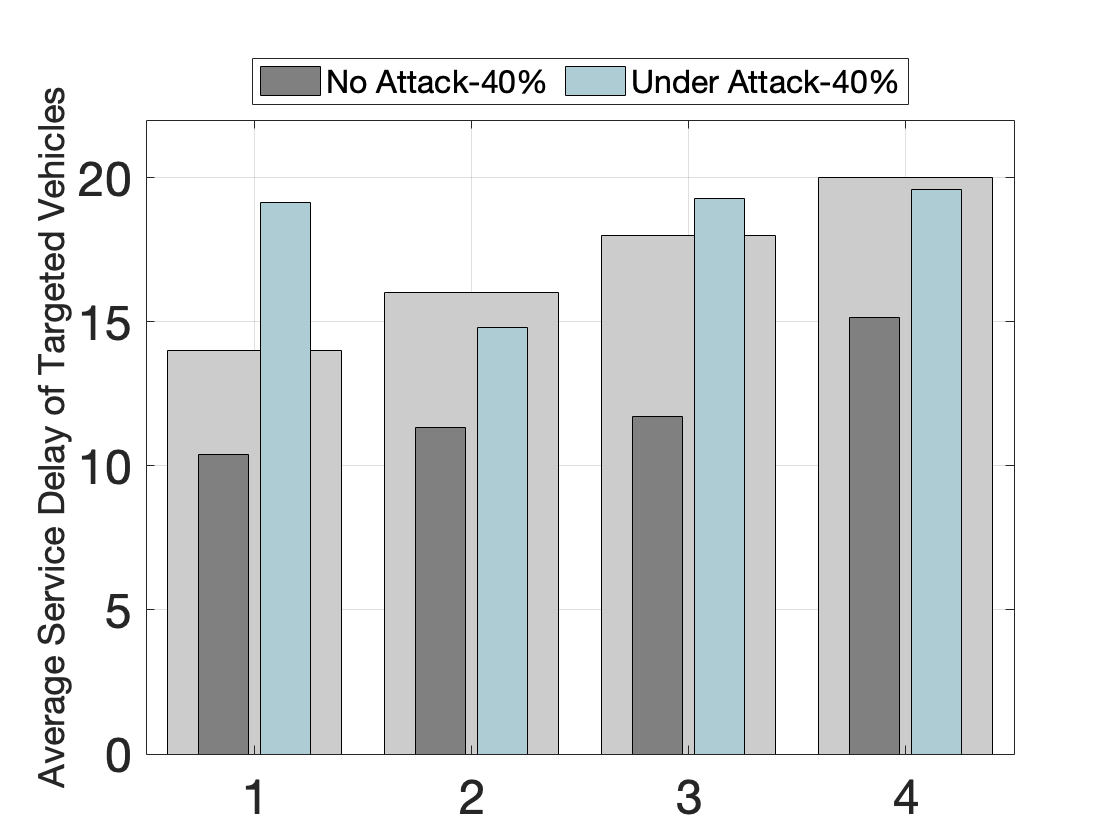}  
		\caption{ }
		\label{fig:2Attdelay40}
	\end{subfigure} 
	\begin{subfigure}{.19\textwidth}
		\centering
		\includegraphics[width=1.5in]{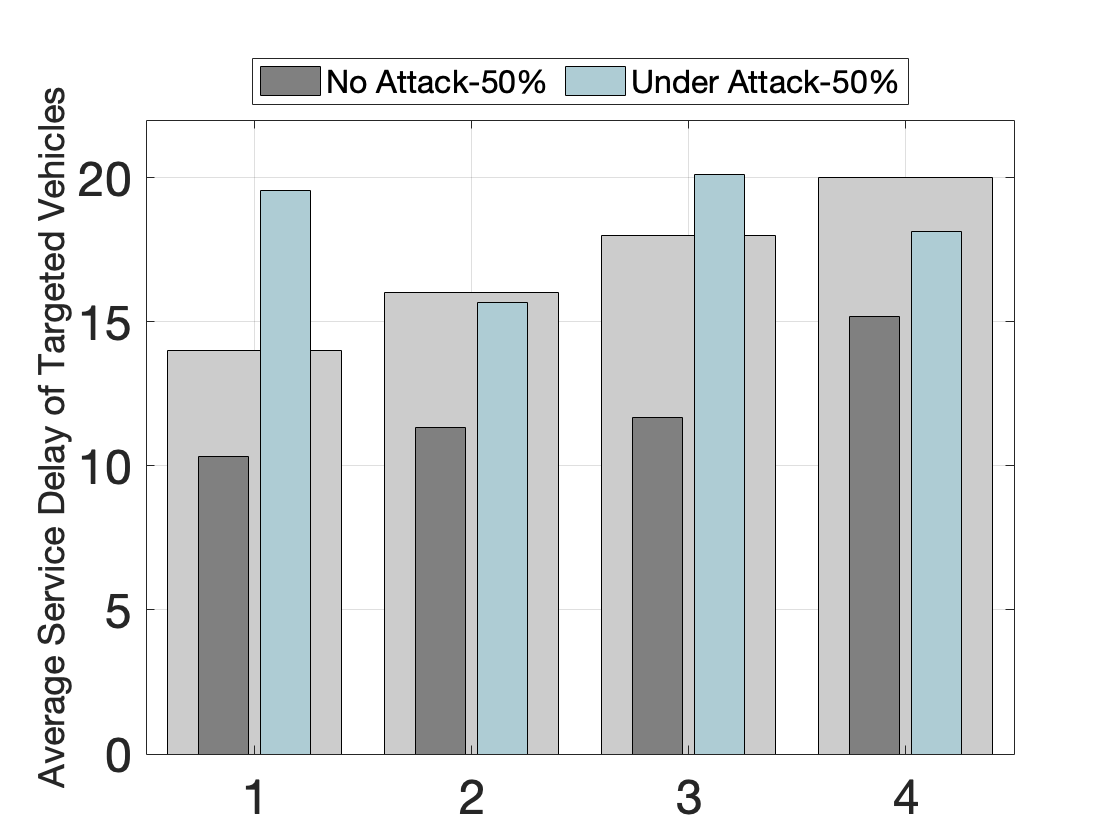}  
		\caption{ }
		\label{fig:2Attdelay50}
	\end{subfigure} 
	\caption{Average Service Delay of Targeted Vehicles - Attack-Selective}
	\label{fig:Attdelay02}
\end{figure*}

\vspace{-5pt} 
Fig. \ref{fig:Avgdelay}, \ref{fig:Attdelay01} and \ref{fig:Attdelay02} depict the average service delay for all vehicles, the targeted vehicles of attack-any case, and the targeted vehicles of attack-selective case, respectively. Here, the targeted vehicles mean the vehicles whose IDs are stolen to create Sybil nodes. This metric calculates the average delay experienced by vehicles for availing different services. Fig. \ref{fig:Avgdelay1} plots the performance for attack-any case where the attacker is amateur and attacks randomly-chosen vehicles without taking note of any further information. In general, the delay is higher for most of the services in an attack scenario than in a no-attack scenario, but the pattern of observed delay is quite random. This is because with insufficient amount of information a random pattern of services is attacked during each time unit. However, even with the little information, an adversary is successful to disrupt the functioning of the DRL algorithm by increasing the average delay for multiple services.\par 
On the contrary, in attack-selective case, when an attacker follows a fixed pattern of services to attack, a significant impact is observed over the delay. The higher proportion of attacked vehicles inevitably increases the average observed delay for all vehicles requesting that service. Additionally, the DRL agent is no more effective to satisfy the delay threshold requirement for the attacked services. In Fig. \ref{fig:Attdelay01} and \ref{fig:Attdelay02}, we plot the delay for the targeted vehicles only. In attack-any case, the delay observed shows an increase of up to 30\% for few services. However, in attack-selective case, the delay for the attack scenario is always higher with an increase of up to 90\% by maliciously reducing the feedback signal of selected services only. Such an increase in delay due to adversarial attacks over DRL can be much serious when the requested service performs critical tasks related to driving.  \par
\vspace{-5pt}

\begin{figure}[hbt!]
	\centering
	\begin{subfigure}{.23\textwidth}
		\centering
		\includegraphics[width=1.8in]{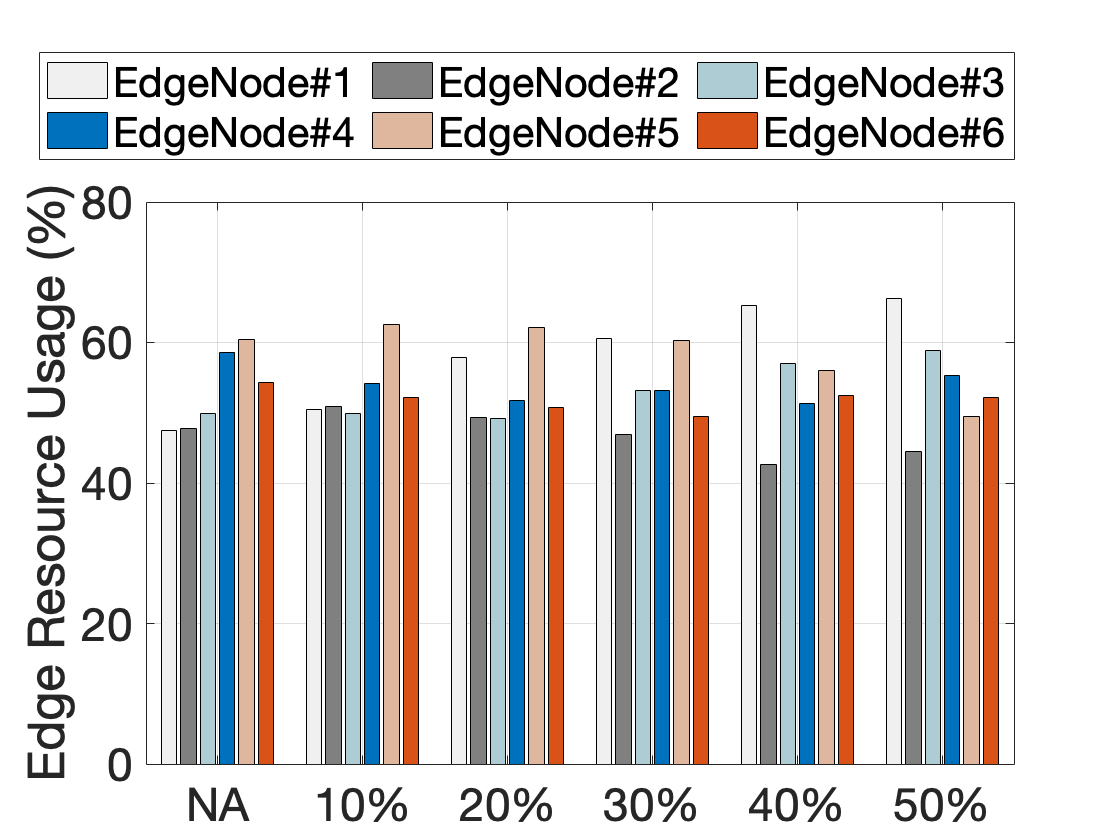}  
		\caption{Attack-Any}
		\label{fig:ERU1}
	\end{subfigure}
	\begin{subfigure}{.23\textwidth}
		\centering
		\includegraphics[width=1.8in]{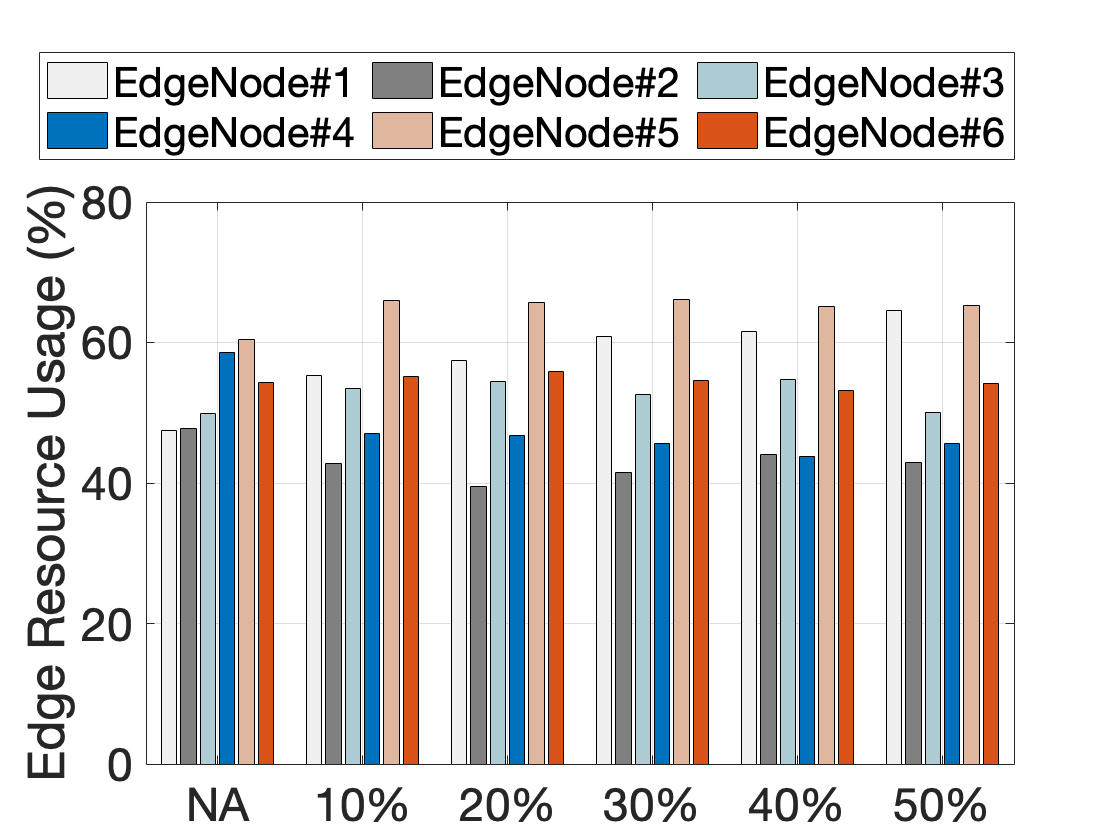}  
		\caption{Attack-Selective}
		\label{fig:ERU2}
	\end{subfigure} 
	\caption{Edge Resource Usage}
	\label{fig:ERU}
\end{figure}
Fig. \ref{fig:ERU} plots the edge resource usage which is measured as the percentage of resources that the instance(s) of services will consume after the placement at the edge node. The resources at the edge servers are limited and thus are important to be utilized efficiently. The impact of adversarial attacks over the DRL framework is much more sensitive for the edge resource usage. With the increasing proportion of attacked vehicles, the imbalanced deployment of resources across the edge nodes increases linearly. Defending the  adversarial attacks over DRL is important because ineffective distribution of edge resources may result in congestion and failure to deliver a service in the case of increasing future demands for the deployed service. \par 
\begin{figure}[hbt!]
	\centering
	\begin{subfigure}{.23\textwidth}
		\centering
		\includegraphics[width=1.8in]{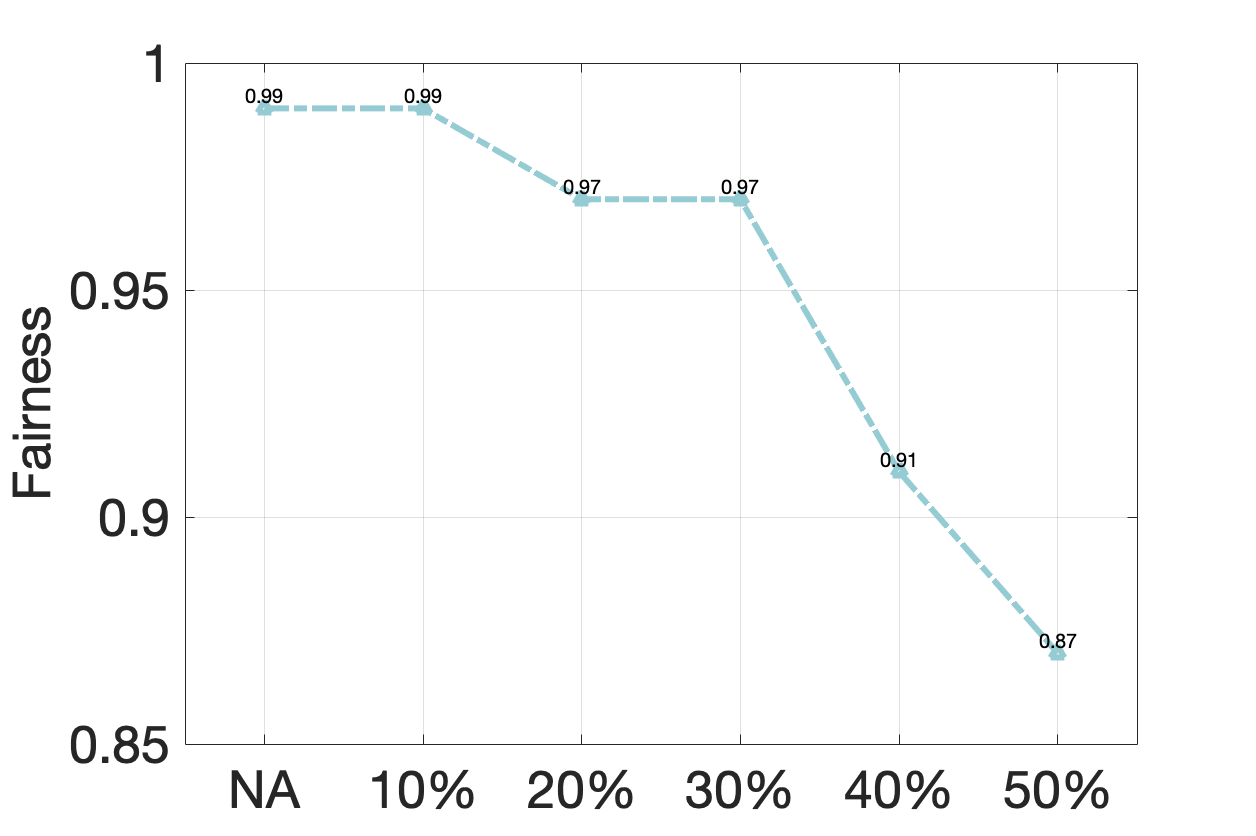}  
		\caption{Attack-Any}
		\label{fig:fairness1}
	\end{subfigure}
	\begin{subfigure}{.23\textwidth}
		\centering
		\includegraphics[width=1.8in]{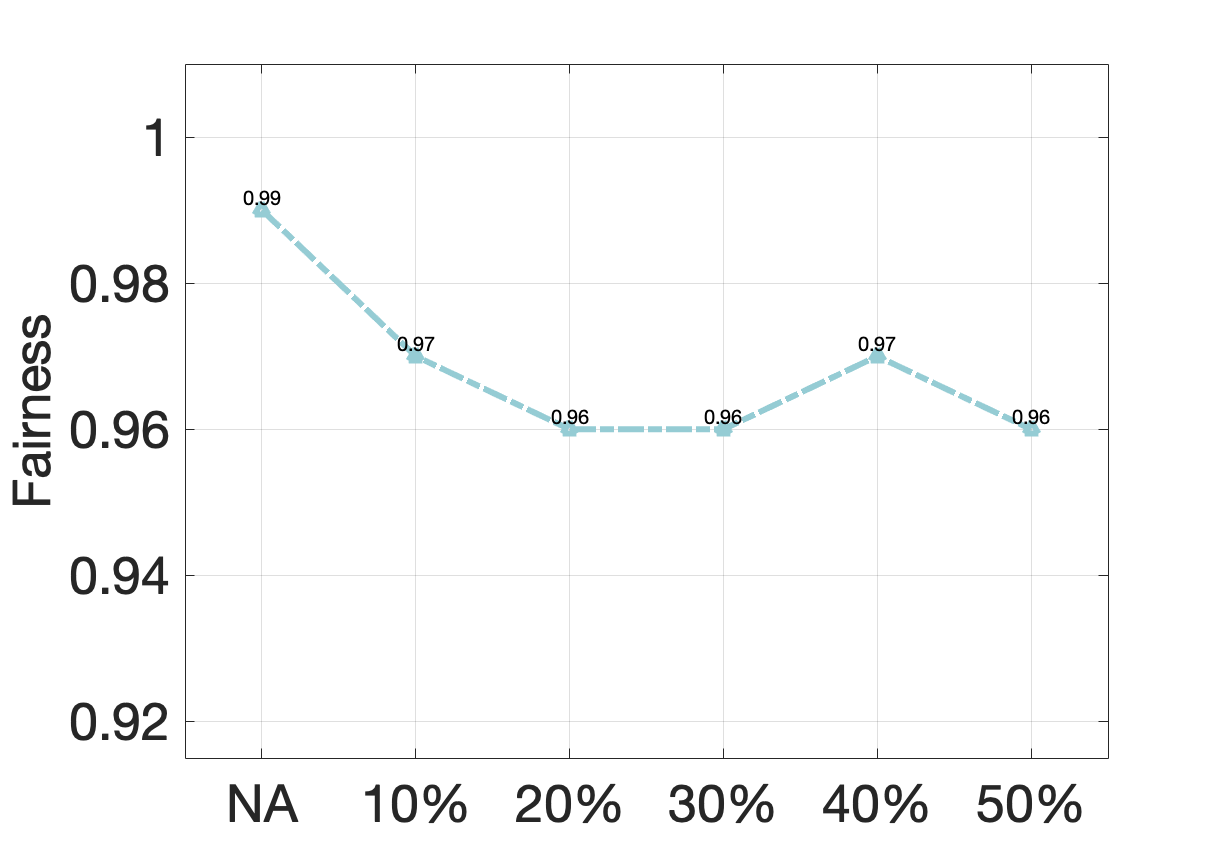}  
		\caption{Attack-Selective}
		\label{fig:fairness2}
	\end{subfigure} 
	\caption{Fairness in Resource Usage}
	\label{fig:fairness}
\end{figure}
To quantitively measure the fair and balanced resource consumption, we plot fairness in Fig, \ref{fig:fairness}. We use Jain's index as a fairness measure in this work \cite{jain_index}. The resource utilization is fair when the index value is higher. As can be observed from Fig. \ref{fig:fairness}, the fairness decreases for both cases as the attack becomes more pronounced with increasing Sybil nodes. In the NA scenario, the fairness is substantially higher compared to all other scenarios. The attack over DRL results in wrong placement decisions which cause significant load imbalance among the edge nodes. This is undesirable, as it may cause bottleneck at any single server given the limited resources of edge.

\section{Conclusion and Future Work}
\label{Sec:conclude}
Security of ML algorithms and their variants is an important and active area of research today. With the focus on the adversarial attack against the DRL framework in IoV, we show that a simple attack can have a significant impact when security measures are not taken into account. In this paper, we studied the problem of adversarial Sybil-based data poisoning attacks against a DRL-based IoV network. Specifically, we considered a DRL-based service placement application to evaluate the impacts on decision-making and network performance in the presence of Sybil nodes. An extensive set of experiments was carried out to show the performance degradation where vehicles experience poor delay and resource congestion when the network is under attack. Such attacks can have devastating effects when the services are performing critical tasks related to vehicle driving. Considering the findings and insights from our study, in the future, we plan to work on developing pre-processing techniques in conjunction with the DRL model to detect or minimize the effect of such adversarial attacks on the DRL framework.

\balance 

\bibliographystyle{IEEEtran}
\bibliography{IEEEabrv,References}

\end{document}